\documentclass[conference,letterpaper]{IEEEtran}
\usepackage[T1]{fontenc}
\usepackage{amsmath,amssymb,bm}
\usepackage{graphicx,booktabs,array,cite,microtype}
\usepackage{algorithm,algpseudocode}
\usepackage{placeins}
\usepackage[hidelinks]{hyperref}
\graphicspath{{figures/}}
\newcommand{\R}{\mathbb{R}}
\newcommand{\norm}[1]{\left\lVert#1\right\rVert}
\newcommand{\pos}[1]{\left[#1\right]_{+}}
\newcommand{\diag}{\operatorname{diag}}
\newcommand{\softmax}{\operatorname{softmax}}

\allowdisplaybreaks
\title{Learning Waypoints and Passage Events for\break Decentralized Multi-Quadrotor Trajectory Planning}
\author{\IEEEauthorblockN{Yunhan Shi\textsuperscript{1}, Jia Deng\textsuperscript{2}, Zehao Jia\textsuperscript{2},\\
Lin Cheng\textsuperscript{1,*}, Shengping Gong\textsuperscript{1}}
\IEEEauthorblockA{\textsuperscript{1}School of Astronautics, Beihang University, Beijing 100191, China\\
\textsuperscript{2}Astroup(Beijing) Technology Co., Ltd.\\
\texttt{syunham@buaa.edu.cn}, \texttt{dengjia@astroup.com}, \texttt{jiazehao@astroup.com}\\
\textsuperscript{*}Corresponding author: \texttt{chenglin5580@buaa.edu.cn}\\
\texttt{gongsp@buaa.edu.cn}}}
\hypersetup{pdftitle={Learning Waypoints and Passage Events for Decentralized Multi-Quadrotor Trajectory Planning},
pdfauthor={Yunhan Shi, Jia Deng, Zehao Jia, Lin Cheng, Shengping Gong}}

\begin{document}
\maketitle

\begin{abstract}
Multi-quadrotor passage planning requires coordinated spatial routes and passage times under vehicle dynamics and actuation constraints. Optimizing these decisions jointly couples temporal allocation, nonlinear motion, and inter-vehicle interactions, creating a computational challenge for local trajectory generation. We propose a learning--optimization framework that predicts spatial waypoints and ordered passage events for independently solved quadrotor planners. A common physical time grid aligns the vehicles, with intermediate events fixed by the networks and terminal progress optimized locally to permit different completion metrics. Differentiable sequential convex programming incorporates rigid-body dynamics and rotor-thrust bounds, allowing collision and completion gradients to train both networks through the generated trajectories. In a two-quadrotor evaluation across 24 layouts, the adapted policy passes physical feasibility and sampled separation checks in all five held-out layouts. Their mean completion metric decreases by 1.52\% relative to the inherited policy. The median forward time is 21.01\,ms on the held-out exchange layout with matching numerical trajectory references. These results demonstrate that learned spatial and event parameters can coordinate independent rigid-body planners while preserving local vehicle constraints in the evaluated scenarios.
\end{abstract}


\section{Introduction}
Multi-quadrotor passage planning concerns the coordinated motion of several vehicles through prescribed gates toward individual targets. Such tasks arise in swarm racing and navigation through cluttered environments, where vehicles share limited passage space~\cite{shen2023,zhou2021ego}. Successful traversal requires each trajectory to satisfy vehicle dynamics and each pair of vehicles to maintain suitable separation. Spatial route selection and passage timing therefore form a common planning problem for coordinated flight.

The central computational difficulty is the coupling between passage timing, vehicle motion, and inter-vehicle separation. Changing an arrival time changes where a vehicle meets its neighbors, and generating the corresponding motion requires satisfying nonlinear attitude and actuation constraints. Solving these decisions jointly introduces temporal variables and interaction constraints into the trajectory optimization. This coupling motivates three complementary approaches: direct optimization, learned motion generation, and hybrid learning--optimization planning.

Optimization-based methods provide explicit representations of trajectory objectives and physical constraints. Minimum-snap and MINCO formulations exploit polynomial structure for efficient trajectory generation~\cite{mellinger2011,wang2022minco}, while complementary progress constraints (CPC) optimize ordered waypoint completion together with motion~\cite{foehn2021,shen2023}. Decentralized planners such as MADER and Robust MADER incorporate shared trajectory information into local collision handling~\cite{tordesillas2022,kondo2024}. The resulting formulations make passage allocation and inter-vehicle coupling explicit, identifying the decisions that a learned planner can help organize.

Learning-based methods offer a direct mapping from observations to motion decisions. Sensor-to-trajectory policies have been trained through privileged imitation learning for high-speed flight~\cite{loquercio2021}, and differentiable simulation has supplied analytical gradients for quadrotor control from states and visual features~\cite{heeg2025}. Their training objectives encode desired behavior in network parameters and support rapid inference. For constrained multi-vehicle passage, this motivates learning spatial and temporal decisions together with a physical trajectory-generation model.

Hybrid learning--optimization methods provide a mechanism for coupling neural predictions to constrained planning. AllocNet predicts segment durations for a differentiable trajectory quadratic program~\cite{wu2024}, Jiang et al. jointly learn waypoints and time allocations~\cite{jiang2025}, and Marino et al. combine learned multi-UAV trajectory parameters with a differentiable correction layer~\cite{marino2024}. These approaches show how optimization gradients can shape planning parameters during training. Inspired by these approaches, we learn spatial and event parameters to coordinate independently solved rigid-body trajectories on a shared time grid.

This design involves three linked challenges. First, pairwise separation must be evaluated at common physical times even when the vehicles complete their routes at different times. Second, the trajectory layer must incorporate nonlinear rigid-body motion and rotor-level limits while retaining usable derivatives for network training. Third, independent local solves need spatial and temporal decisions that respond to neighboring vehicles. A suitable hierarchy must therefore connect synchronized event allocation, constrained local motion, and inter-vehicle learning within one training objective.

\begin{figure*}[!t]
\centering
\includegraphics[width=\textwidth]{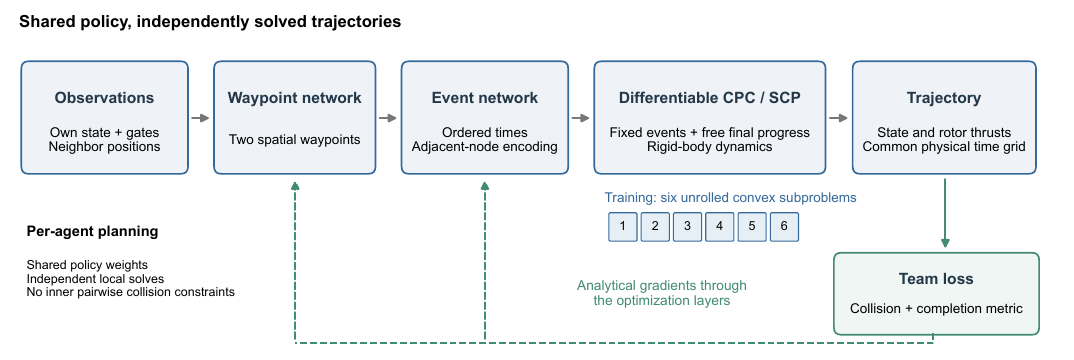}
\caption{Schematic of the learning and planning architecture. The waypoint block illustrates two learned intermediate points, which are combined with prescribed gate centers to define the passage events. The event block assigns their temporal locations on a shared grid. The repeated SCP blocks illustrate differentiable local planning under linearized rigid-body dynamics. Dashed green arrows denote gradients of the outer collision and completion objectives. The waypoint count and unrolling depth shown here are example design choices.}
\label{fig:architecture}
\end{figure*}

We propose a learning--optimization framework that connects these three requirements through spatial waypoints and passage events. As illustrated in Fig.~\ref{fig:architecture}, shared networks use the vehicle state, known gate locations, and current relative neighbor positions to predict waypoint centers and ordered intermediate events. Independent differentiable CPC-based planners then generate trajectories on a common physical grid, and the outer objective evaluates collision and completion costs across the resulting trajectories. This hierarchy makes learned event allocation the connection between multi-vehicle coordination and local physical planning.

The framework has three main technical contributions:
\begin{enumerate}
\item A shared-grid event representation organizes spatial targets and passage timing for independent local planners. Continuous intermediate timestamps parameterize interpolated passage constraints, and the final CPC progress distribution remains a local decision variable. This construction supports common-time interaction evaluation together with vehicle-specific completion metrics.
\item A differentiable sequential convex programming (SCP) trajectory layer incorporates rigid-body dynamics and individual rotor-thrust limits. Local approximations of the dynamics and terminal progress yield convex conic subproblems whose parameterization satisfies disciplined parametrized programming. The layer provides a constrained trajectory map through which waypoint and event predictions can be trained.
\item An outer coordination objective adapts both networks through independently generated trajectories. Collision penalties act on synchronized vehicle positions, and the completion term links the learned parameters to local progress. This objective makes inter-vehicle behavior part of parameter learning while keeping the trajectory solves independent.
\end{enumerate}

The numerical evaluation examines coordination, physical feasibility, and computation within a two-vehicle setting. Gate-perturbation and goal-exchange layouts test spatial and temporal adaptation, a frozen-policy experiment examines thrust-difference regularization, and selected finite differences check the network derivatives. Warm-reference forward timings characterize the solver's computational cost. Together, these experiments assess the framework through trajectory behavior, local constraint satisfaction, and measured planning cost.

\section{Related Work}
\subsection{Waypoint Timing and Trajectory Optimization}
Polynomial minimum-snap trajectories provide an efficient representation for quadrotor flight through waypoint sequences~\cite{mellinger2011}. MINCO provides a compact polynomial representation with efficient spatial--temporal deformation and supports geometrical and dynamic constraints through trajectory optimization~\cite{wang2022minco}. CPC models ordered completion through continuous progress variables and jointly optimizes passage allocation and vehicle motion~\cite{foehn2021}. Shen et al. extend this principle to a swarm racing task~\cite{shen2023}. Our formulation retains complementary progress only for the final event and uses learned continuous timestamps for intermediate events, which restrict the feasible trajectories to those compatible with the assigned passage times.

\subsection{Decentralized Multi-Vehicle Planning}
MADER generates trajectories asynchronously using enclosing polyhedra, optimized separating planes, and collision checks against other agents' committed trajectories~\cite{tordesillas2022}. Robust MADER extends trajectory sharing with delay checks and a two-step publication scheme to accommodate communication delays in dynamic environments~\cite{kondo2024}. EGO-Swarm combines gradient-based local planning, collision-risk penalties, and lightweight topological trajectory generation for decentralized navigation in cluttered environments~\cite{zhou2021ego}. These methods make exchanged trajectory information part of decentralized collision handling. In our formulation, current relative neighbor positions condition the shared networks, and synchronized local trajectories enter the outer collision objective during training.

\subsection{Learning with Differentiable Planners}
Differentiable MPC learns controller costs and dynamics by differentiating the KKT conditions of a convex approximation at a controller fixed point~\cite{amos2018mpc}. AllocNet predicts segment durations and differentiates through the fixed-duration trajectory QP~\cite{wu2024}. Jiang et al. combine depth-based waypoint prediction, time allocation, and differentiable minimum-snap optimization~\cite{jiang2025}. Marino et al. predict MINVO control points and a learned collision constraint from point clouds and communicated features, then use an OptNet-based QP layer during imitation learning~\cite{marino2024,amos2017}. These works establish the broader learning--optimization architecture. Our focus is the event representation and the separation of outer inter-agent collision learning from local rotor-level planning.

For interacting aerial systems, learned model predictive games project neural reference strategies into dynamically feasible trajectories and train with game objectives~\cite{papuc2026}. DiffCoord differentiates a distributed ADMM--DDP pipeline to learn problem and solver parameters for cooperative trajectories~\cite{wang2026}. Our local optimizers interact through shared observations and the outer training objective. This choice reduces explicit coupling in each local solve, while leaving collision avoidance dependent on the learned policy and numerical validation.

Learning also connects visual observations to aerial motion. Loquercio et al. train a sensor-to-trajectory policy through privileged imitation learning in simulation for high-speed flight~\cite{loquercio2021}. Heeg et al. train quadrotor control policies using analytical gradients through differentiable dynamics and learned visual-state representations~\cite{heeg2025}. These studies provide context for trajectory-level learning and dynamics-based policy gradients. The present formulation learns spatial and event parameters from state observations through differentiable local optimization.

\section{Problem Formulation}
We consider a multi-quadrotor racing scenario in which vehicles traverse prescribed gates and proceed to individual terminal targets. Each quadrotor has real-time access to its own state and the race-gate positions, and observes the current relative positions of neighboring quadrotors. These local observations provide the spatial information used to coordinate passage timing and generate dynamically feasible trajectories.

\subsection{Observations and Shared Time Grid}
Consider $n_a$ vehicles indexed by $i$. The local observation is
\begin{equation}
 o_i=\left(x_{i,0},g_{i,1},g_{i,2},
             \{p_{j,0}-p_{i,0}\}_{j\in\mathcal N_i}\right),
\end{equation}
where $\mathcal N_i$ is the set of neighboring vehicles observed at the current planning instant, $g_{i,1}$ is the shared passage center, and $g_{i,2}$ is the individual terminal target. Neighbor observations consist of current relative positions. The formulation is state based.

Every vehicle uses the same grid $t_k=kh$, $k=0,\ldots,N$, with a common horizon $H=Nh$. Independent completion events may occur at different grid locations. Full trajectories are retained to $H$, including the segments after the completion event, and remain active in collision evaluation.

\subsection{Rigid-Body Dynamics}
The state and input are
\begin{equation}
 \begin{aligned}
 x_i&=(p_i,v_i,q_i,\omega_i)\in\R^{13},\\
 u_i&=(f_{i,0},f_{i,1},f_{i,2},f_{i,3})\in\R^4,
 \end{aligned}
\end{equation}
where $q_i$ is a scalar-first quaternion and $f_{i,r}$ is an individual rotor thrust. With $e_3=(0,0,1)^\top$, the model is
\begin{equation}
 \begin{aligned}
 \dot p_i&=v_i,\\
 \dot v_i&=m^{-1}\Big(\sum_r f_{i,r}\Big)R(q_i)e_3-ge_3,\\
 \dot q_i&=\tfrac12 q_i\otimes(0,\omega_i),\\
 \dot\omega_i&=J^{-1}\left(\tau_i-\omega_i\times J\omega_i\right).
 \end{aligned}
 \label{eq:dynamics}
\end{equation}
Rotor thrusts map to torque through the X-configuration allocation
\begin{equation}
 \tau_i=
 \begin{bmatrix}
 a&a&-a&-a\\-a&a&a&-a\\c_\tau&-c_\tau&c_\tau&-c_\tau
 \end{bmatrix}u_i,\qquad a=\ell/\sqrt{2}.
\end{equation}
The discrete transition map $F_h$ propagates the rigid-body state over each time interval with a unit-quaternion attitude representation. The model resolves rotor thrusts, attitude, and angular-rate dynamics under a rigid-body assumption with gravity and rotor forces.

\subsection{Passage Events and Coordination Objective}
Each vehicle follows $E$ ordered event centers that interleave learned spatial waypoints and prescribed gates, with the terminal target as the last event. For example, the two-waypoint schematic in Fig.~\ref{fig:architecture} corresponds to
\begin{equation}
 y_{i,1:4}=(w_{i,1},g_{i,1},w_{i,2},g_{i,2}),
\end{equation}
where $w_{i,1:2}$ denote learned intermediate points, $g_{i,1}$ is the shared passage center, and $g_{i,2}$ is the terminal target. Each event couples a spatial center, a tolerance radius, and a passage-time condition. A gate is represented by a position tolerance ball $\norm{p-y_{i,j}}_2\le r_{i,j}$, which specifies positional proximity to its center.

Training minimizes an objective of the form
\begin{equation}
 \min_\theta\ \mathbb{E}_{s\in\mathcal D_{\rm train}}
 \left[\mathcal L_{\rm col}(X^\theta)
       +\eta\frac{1}{n_aH}\sum_i T_i^\theta\right],
 \label{eq:outer}
\end{equation}
where $(X_i^\theta,U_i^\theta,\mu_{i,E}^\theta)$ are outputs of the local planner conditioned on the network predictions. $T_i$, defined in Sec.~\ref{sec:terminal}, is an interval-index completion metric.

\section{Learned Spatial and Event Parameters}
\subsection{Waypoint Network}
A shared encoder maps each relative neighbor position to a feature vector using a linear layer and a $\tanh$ nonlinearity. Mean pooling gives a fixed-size neighborhood embedding, which is concatenated with the own state and gate coordinates. A waypoint head predicts bounded offsets from a smooth geometric reference. For example, the waypoint block in Fig.~\ref{fig:architecture} represents intermediate spatial points whose locations can adjust the route before and after a prescribed gate. These points become event centers for the local planner; their passage times are assigned by the event network.

\subsection{Ordered Event Encoding}
The event network receives the ordered event centers and radii. Its hidden layer produces corrections to $E+1$ baseline gap logits, including the gaps before the first and after the last event. For $N>E+1$, with $z_i\in\R^{E+1}$ denoting corrected logits, define
\begin{align}
 d_{i,j}&=1+(N-E-1)\softmax(z_i)_j,\label{eq:gaps}\\
 c_{i,j}&=\sum_{l=1}^{j}d_{i,l},\quad j=1,\ldots,E.
 \label{eq:centers}
\end{align}
Thus the gaps sum to $N$, each exceeds one grid interval, and the event centers remain ordered inside the horizon.

\begin{figure}[!t]
\centering
\includegraphics[width=\columnwidth]{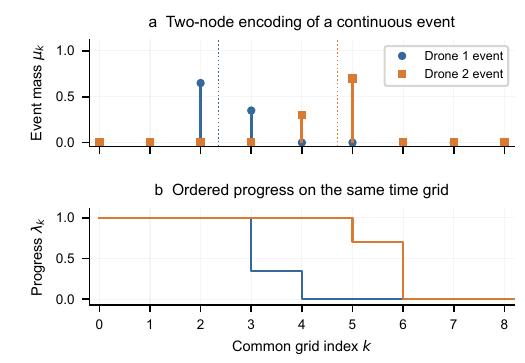}
\caption{Schematic of event encoding on a shared time grid. (a) Each color illustrates one vehicle's event: stem heights are the adjacent-node masses $\mu^{\rm enc}$, and the dotted vertical line marks their grid-index expectation $c$. (b) The associated progress $\lambda$ decreases from one to zero. The event locations illustrate the representation. Each intermediate code parameterizes a single continuous-time passage constraint at its fractional timestamp.}
\label{fig:event}
\end{figure}

For $k_j=\lfloor c_{i,j}\rfloor$ and $\alpha_j=c_{i,j}-k_j$, the adjacent-node code is
\begin{equation}
 \mu^{\rm enc}_{i,j,k}=
 (1-\alpha_j)\mathbf{1}_{k=k_j}
 +\alpha_j\mathbf{1}_{k=k_j+1}.
 \label{eq:encoding}
\end{equation}
Its mass sums to one and its grid-index expectation is $c_{i,j}$. For example, the colored stems in Fig.~\ref{fig:event}(a) represent the masses assigned to two adjacent nodes for an illustrative event of each vehicle. The dotted vertical lines mark the corresponding fractional event locations. Panel~(b) shows the associated remaining progress, which decreases as the event is completed. This encoding retains gradients inside a selected grid cell. The cell-selection operation is discrete, so gradients are evaluated locally within the selected event support.

\subsection{Continuous Intermediate Event Constraints}
The first $E-1$ event times $\tau_{i,j}=hc_{i,j}$ are fixed during each local solve. At $\alpha\in[0,1]$ in cell $k$, the event position is interpolated from the adjacent position and velocity states as
\begin{align}
 \hat p_i(\tau)={}&H_{00}(\alpha)p_{i,k}
     +hH_{10}(\alpha)v_{i,k}\nonumber\\
 &+H_{01}(\alpha)p_{i,k+1}
     +hH_{11}(\alpha)v_{i,k+1},\label{eq:event-interpolation}
\end{align}
where $H_{00}=2\alpha^3-3\alpha^2+1$, $H_{10}=\alpha^3-2\alpha^2+\alpha$, $H_{01}=-2\alpha^3+3\alpha^2$, and $H_{11}=\alpha^3-\alpha^2$. The event constraint is
\begin{equation}
 \norm{\hat p_i(\tau_{i,j})-y_{i,j}}_2
 \le r_{i,j}-\delta_r,\quad j=1,\ldots,E-1,
 \label{eq:fixed-event}
\end{equation}
where $\delta_r>0$ is an interpolation margin. For fixed event coefficients, this is a second-order-cone constraint affine in the state variables. The margin compensates numerically for interpolation discrepancy; an independent dynamics rollout checks the integrated motion at the assigned event times.

\subsection{Free Terminal Progress}\label{sec:terminal}
The last event retains nonnegative decision variables $\mu_{i,E,k}$. Progress satisfies
\begin{equation}
 \lambda_{i,j,0}=1,\quad
 \lambda_{i,j,k+1}=\lambda_{i,j,k}-\mu_{i,j,k},\quad
 \lambda_{i,j,N}=0.
\end{equation}
Intermediate rows use~\eqref{eq:encoding}; the final row is optimized. Define $C_{i,j,k}=\sum_{l=0}^{k}\mu_{i,j,l}$. The condition $C_{i,E,k}\le C_{i,E-1,k}$ prevents terminal progress from preceding completion of the previous event. Terminal mass is normalized and restricted to support beginning at the preceding event cell. The remaining nonredundant cumulative-order inequalities are imposed explicitly.

After eliminating the original CPC tolerance variable, the relaxed terminal condition is
\begin{equation}
 \mu_{i,E,k}\left(\norm{p_{i,k}-g_{i,2}}_2^2-r_{i,E}^2\right)
 \le\varepsilon,
 \label{eq:terminal-product}
\end{equation}
where $\varepsilon>0$ is the complementarity relaxation tolerance. The interval-index completion metric is
\begin{equation}
 T_i=h\sum_{k=0}^{N-1}(k+1)\mu_{i,E,k}.
 \label{eq:completion}
\end{equation}
The $(k+1)$ convention associates terminal progress with the end of an indexed interval. The distance in~\eqref{eq:terminal-product} is evaluated at node $k$. Thus $T_i$ measures the expected interval-end index of terminal progress. The local optimizer chooses the final mass distribution subject to the terminal proximity and ordering constraints.

\section{Differentiable Local Planning and Training}
\subsection{Sequential Convex Subproblems}
SCP constructs local convex trajectory subproblems using dynamics approximations and trust-region updates~\cite{malyuta2022}. At iteration $s$, linearization of the discrete transition map $F_h$ around $(\bar x_k,\bar u_k)$ gives
\begin{align}
 x_{k+1}&=A_k^{(s)}x_k+B_k^{(s)}u_k+b_k^{(s)},\\
 A_k^{(s)}&=\partial_x F_h(\bar x_k,\bar u_k),\qquad
 B_k^{(s)}=\partial_u F_h(\bar x_k,\bar u_k).
\end{align}
The affine offset reproduces $F_h$ at the reference. Quaternion normalization is included in these Jacobians.

For the terminal product, let $\bar d_k=\norm{\bar p_k-g_{i,2}}_2^2$ and $a_k=\max(\bar\mu_k,a_{\min})$, with a positive curvature floor $a_{\min}$. The convex partial-product approximation is
\begin{equation}
 a_k\norm{p_k-g_{i,2}}_2^2
 +(\bar d_k-r_{i,E}^2)\mu_k-a_k\bar d_k
 \le\varepsilon.
 \label{eq:surrogate}
\end{equation}
Equation~\eqref{eq:surrogate} is a local partial-product approximation to~\eqref{eq:terminal-product}. State, thrust, and progress trust regions restrict each update, and the original nonlinear residual is checked after the solve.

Each convex subproblem minimizes
\begin{align}
 J_i^{(s)}={}&T_i+\rho_u\sum_k\norm{u_k-u_{\rm hov}}_2^2
 +\frac{\rho_\Delta}{h}\sum_{k=0}^{N-2}\norm{u_{k+1}-u_k}_2^2
 \nonumber\\
 &+\rho_x\norm{X-\bar X}_F^2+\rho_0\norm{X}_F^2,
 \label{eq:inner-objective}
\end{align}
subject to the affine dynamics, event constraints, terminal progress conditions, and vehicle bounds. Here $u_{\rm hov}=(mg/4)\mathbf{1}$. The terminal state has $p_N=g_{i,2}$, $v_N=0$, zero quaternion vector part, and $\omega_N=0$.

Speed and angular-rate limits, event balls, and the quadratic terminal surrogate yield convex conic subproblems with second-order-cone constraints. Coefficients such as $\sqrt{a_k}$, $\bar d_k-r_{i,E}^2$, and $-a_k\bar d_k$ are formed outside CVXPY~\cite{diamond2016} and passed as separate parameters. Auxiliary residual variables separate parameterized targets from quadratic expressions. This construction permits a disciplined parametrized programming (DPP) template embedded with CvxpyLayer~\cite{agrawal2019layers}.

\subsection{Training Objective and Gradients}
At sampled physical times $t\in\mathcal T$, define
\begin{equation}
 D_{ij}(t)=\norm{S(p_i(t)-p_j(t))}_2,\qquad
 S=\diag(s_x,s_y,s_z).
\end{equation}
The collision loss is
\begin{equation}
 \mathcal L_{\rm col}
 =\frac{1}{|\mathcal P|\,|\mathcal T|}
 \sum_{(i,j)\in\mathcal P}\sum_{t\in\mathcal T}
       \pos{d_{\rm train}-D_{ij}(t)}^2,
 \label{eq:collision-loss}
\end{equation}
where $\mathcal P$ is the set of unordered vehicle pairs and the positive diagonal entries of $S$ set the directional distance scaling. The training margin $d_{\rm train}$ and evaluation threshold $d_{\rm eval}$ both specify required inter-vehicle distances in this scaled metric.

The analytical chain from the outer loss to waypoint and event parameters passes through $K$ unrolled SCP steps. Each CvxpyLayer differentiates its conic solution map using diffcp~\cite{agrawal2019cone}. Dynamics derivatives and event interpolation coefficients remain in the computational graph, which enables optimization-based gradient training of both networks.

\begin{algorithm}[t]
\caption{Training with independent event-conditioned planners}
\label{alg:train}
\begin{algorithmic}[1]
\Require Training layouts, shared network weights $\theta$, numerical references
\For{each training update}
  \State Sample a batch of training layouts
  \For{each layout and each vehicle $i$}
    \State Predict spatial waypoints and ordered intermediate events
    \State Construct event interpolation coefficients and terminal support
    \For{$s=1,\ldots,K$}
      \State Linearize $F_h$ and form terminal coefficients
      \State Solve the DPP-safe convex conic subproblem
      \State Update state, thrust, and terminal-progress reference
    \EndFor
  \EndFor
  \State Evaluate~\eqref{eq:outer} on synchronized trajectories
  \State Backpropagate and propose an Adam update
  \State Accept a feasible loss-decreasing candidate using backtracking
\EndFor
\end{algorithmic}
\end{algorithm}

\subsection{Numerical Forward Backend}
The forward backend reuses the sparse conic template and directly updates Clarabel numerical data~\cite{goulart2024}. A cached dynamics function supplies the state and control derivatives used for linearization~\cite{andersson2019}. Adaptive stopping checks the current nonlinear dynamics defect, terminal CPC residual, physical bounds, event constraints, and quaternion consistency. A stored trajectory can initialize the numerical reference when available; otherwise, the planner constructs an initial reference for the requested layout.

Template reuse and numerical-reference reuse are separate mechanisms: the former preserves the problem structure, while the latter also supplies state, thrust, and progress values. Their evaluation protocols are specified in Sec.~VI. No online pairwise collision constraint is added to the local planner.

\section{Experimental Evaluation}
\subsection{Setup and Evaluation Protocol}
Table~\ref{tab:parameters} lists the final configuration. The two vehicles start at $(-0.5,-0.3,1)$\,m and $(-0.5,0.3,1)$\,m, with zero velocity, identity attitude, and zero body rate. Twenty original gate-perturbation layouts retain their split of 12 training, four validation, and four test layouts. Four additional layouts exchange the destination sides across a shared gate: two are used for training, one for validation, and one for held-out testing. The combined split is therefore 14/5/5.

The implementation uses two learned waypoints and four event centers, as illustrated in Fig.~\ref{fig:architecture}, with $h=0.15$\,s, $N=20$, and $H=3$\,s. The neighbor encoder has 16 features; the own state and gate coordinates give a 35-dimensional concatenated input. Both network heads use 48 hidden units. The waypoint head produces six offsets, bounded componentwise by $(0.10,0.25,0.05)$\,m, around reference points at grid indices 5 and 15. The event head corrects five baseline gap logits.

Training unrolls $K=6$ convex subproblems, with three integration substeps per interval and conic-solver differentiation through diffcp. The interpolation margin is $\delta_r=0.0003$\,m, and $\varepsilon=a_{\min}=10^{-5}$. Neither virtual dynamics controls nor terminal CPC slack variables are used. The final progress is initialized by shifting the preceding event row by one interval. The CVXPY template passes DPP checks. Each training update samples one exchange layout and one ordinary layout. The collision loss uses five subinterval samples per interval, $S=\diag(1,1,1/3)$, and $\eta=0.005$ in~\eqref{eq:outer}.

\begin{table}[!t]
\caption{Rigid-body planning and training configuration}
\label{tab:parameters}
\centering\small
\begin{tabular}{@{}ll@{}}
\toprule
Parameter & Value\\\midrule
Vehicle mass $m$ & 0.713\,kg\\
Inertia diagonal $J$ & $(0.03,0.03,0.055)$\,kg\,m$^2$\\
Arm length $\ell$; torque coefficient $c_\tau$ & 0.17\,m; 0.01\,m\\
Rotor thrust range & $[1.0,3.3]$\,N\\
Speed; body-rate limit & 1.5\,m/s; 8\,rad/s\\
Height range & $[0.5,1.5]$\,m\\
Grid step; intervals; training SCP steps & 0.15\,s; 20; 6\\
Waypoint radius & 0.045\,m\\
Shared gate; terminal target radius & 0.24\,m; 0.12\,m\\
Training; evaluation scaled separation & 0.30\,m; 0.25\,m\\
$\rho_u$; $\rho_\Delta$; $\rho_x$; $\rho_0$ & $10^{-3}$; 0.002; $10^{-3}$; $10^{-5}$\\
Waypoint; event learning rate & 0.001; 0.002\\
\bottomrule
\end{tabular}
\end{table}

The exchange adaptation starts from the inherited two-waypoint policy. Both networks remain trainable. Thirty-six candidate updates are accepted. Checkpoints are selected by validation loss; stopping occurs after the validation checkpoints at updates 30 and 36 both pass feasibility and sampled separation checks, with relative loss change below 1\%. Gradient updates use training layouts, and the stopping rule uses validation layouts. The recorded training, validation, and final evaluation loop takes 27.36\,min, excluding preparation and the initial baseline evaluation.

The independent audit rolls out the full rigid-body model with 20 substeps per interval, producing 401 state samples per vehicle over the full horizon. It checks dynamics defects, rotor and state bounds, quaternion consistency, assigned intermediate events, terminal residuals, event order, and pairwise separation. The dynamics-defect tolerance is $5\times10^{-5}$, and the maximum allowed positional mismatch between node states and rollout is 0.002\,m. All attempted layouts are retained in the reported counts. Collision assessment is limited to these sampled times.

\subsection{Multi-Layout Adaptation}
\begin{figure*}[!t]
\centering
\includegraphics[width=\textwidth]{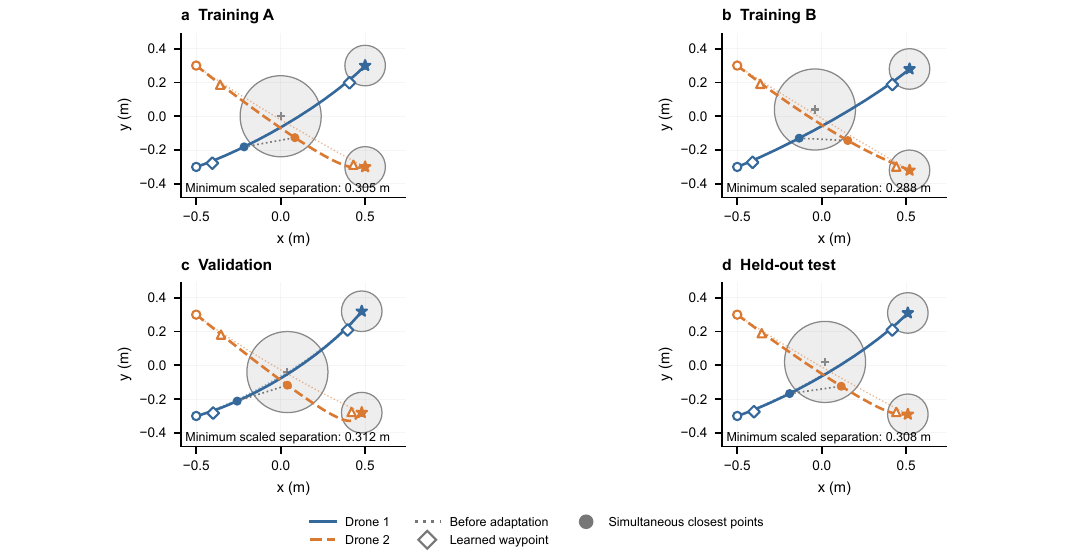}
\caption{Top-view projections of all four added goal-exchange layouts: two training layouts, one validation layout, and one held-out test layout. Solid blue and dashed orange lines are the adapted trajectories; faint dotted lines show the inherited-policy trajectories. Open waypoint markers denote learned spatial targets, circles indicate initial positions, and stars indicate final positions. Gray disks are projections of the position tolerance balls used to represent gates. Filled markers and their dotted connector identify both vehicles at the same closest-approach time. All 401 rollout samples per vehicle are used.}
\label{fig:trajectories}
\end{figure*}

Table~\ref{tab:splits} reports the combined layout split. The inherited policy is physically feasible in all 24 layouts, but the two exchange training layouts, the exchange validation layout, and the exchange test layout violate the separation threshold. After adaptation, all 24 layouts pass both checks. The mean test completion metric decreases from 2.40039 to 2.36384\,s, a 1.52\% reduction relative to the inherited policy, which measures adaptation within the current architecture. Figure~\ref{fig:trajectories} shows the associated exchange trajectories.

\begin{table}[!t]
\caption{Layout-level results before and after exchange adaptation}
\label{tab:splits}
\centering\small
\begin{tabular}{@{}lrrrr@{}}
\toprule
 & \multicolumn{2}{c}{Sampled separation passed} & \multicolumn{2}{c}{Mean $T_i$ (s)}\\
Split & Before & After & Before & After\\\midrule
Training & 12/14 & 14/14 & 2.40039 & 2.36335\\
Validation & 4/5 & 5/5 & 2.40040 & 2.36397\\
Held-out test & 4/5 & 5/5 & 2.40039 & 2.36384\\
\bottomrule
\end{tabular}
\end{table}

\begin{figure}[!t]
\centering
\includegraphics[width=\columnwidth]{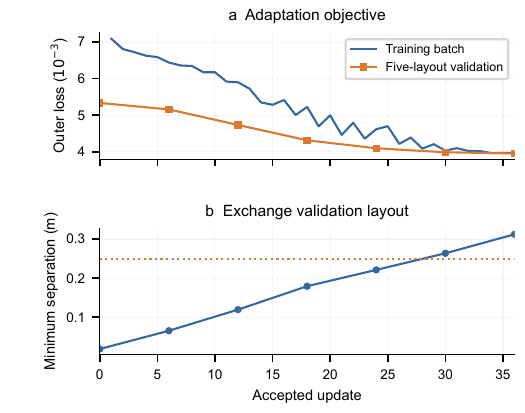}
\caption{Recorded exchange adaptation. (a) All 36 accepted batch updates and seven validation checkpoints, including update zero. Each batch contains one exchange and one ordinary training layout; validation uses five layouts. (b) Minimum audited scaled separation on the single exchange validation layout. The dotted line is the 0.25\,m evaluation threshold. No smoothing or independent-seed confidence interval is applied.}
\label{fig:learning}
\end{figure}

Figure~\ref{fig:learning} shows the recorded batch objective and validation progress. The batch curve reflects changing sampled layouts, while each validation checkpoint covers the same five layouts. The exchange validation clearance increases from approximately 0.019 to 0.312\,m. These curves describe successive updates within one adaptation run.

\subsection{Goal-Exchange Trajectories}
\begin{figure*}[!t]
\centering
\includegraphics[width=\textwidth]{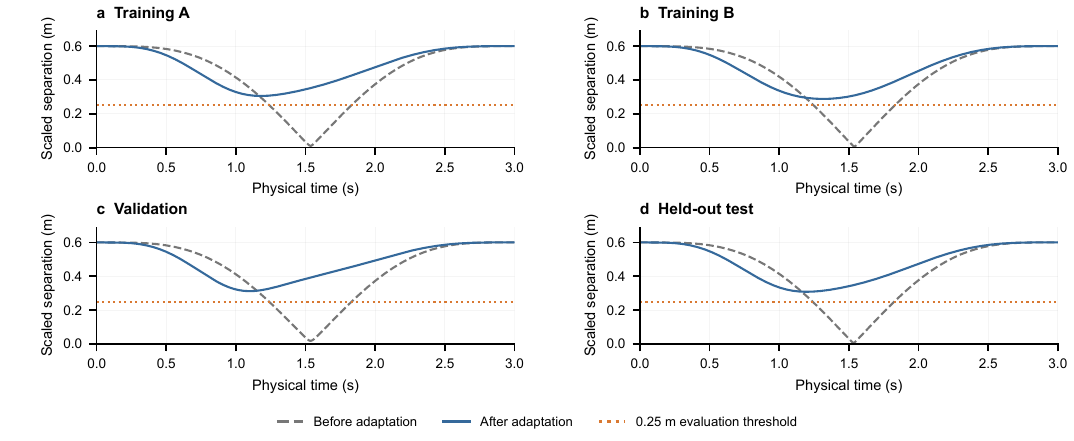}
\caption{Audited scaled inter-vehicle separation over the complete common 3\,s horizon, including post-event trajectory tails. All four exchange layouts are shown. The metric is $\norm{\diag(1,1,1/3)(p_1-p_2)}_2$. The 0.25\,m threshold is exceeded after adaptation in each layout. The panels describe two training layouts, one validation layout, and one held-out test layout.}
\label{fig:clearance}
\end{figure*}

Figure~\ref{fig:trajectories} shows all four added layouts and both learned waypoints per vehicle. Each vehicle finishes on the opposite lateral side from its start. Figure~\ref{fig:clearance} evaluates the vehicles at equal physical times, avoiding the ambiguity of geometric path intersections. The final minimum scaled separation ranges from 0.28798 to 0.31195\,m, compared with 0.00724--0.01877\,m before adaptation (Table~\ref{tab:crossing}). The original 20 layouts continue to satisfy the evaluation threshold.

\begin{table}[!t]
\caption{Added exchange layouts and their temporal separation}
\label{tab:crossing}
\centering\small
\begin{tabular}{@{}lrrr@{}}
\toprule
Layout & \multicolumn{2}{c}{Min. scaled distance (m)} & $\Delta t_{x}$ (s)\\
 & Before & After & After\\\midrule
Training A & 0.01128 & 0.30550 & 0.40319\\
Training B & 0.00737 & 0.28798 & 0.37837\\
Validation & 0.01877 & 0.31195 & 0.42926\\
Held-out test & 0.00724 & 0.30822 & 0.40033\\
\bottomrule
\end{tabular}
\end{table}

The assigned shared-gate event times differ by approximately 0.101\,s after adaptation. The diagnostic first left-to-right crossing of the plane $x=g_{1,x}$ has a time difference $\Delta t_x$ of 0.378--0.429\,s. These plane-crossing times are interpolated from the audited rollout for post-processing. At the closest approaches, horizontal separation is approximately 0.288--0.312\,m, and vertical separation is 0.006--0.018\,m. The observed avoidance is predominantly associated with horizontal and temporal separation under joint adaptation of the waypoint and event networks.

\subsection{Thrust-Difference Regularization}
A separate experiment isolates the $\rho_\Delta$ term with the earlier two-waypoint policy frozen. Both variants use the original 20 layouts (12/4/4 split), the same references, and the same recovery rule; only the inner thrust-difference weight changes. The evaluation uses this experiment's original configuration.

\begin{figure}[!t]
\centering
\includegraphics[width=\columnwidth]{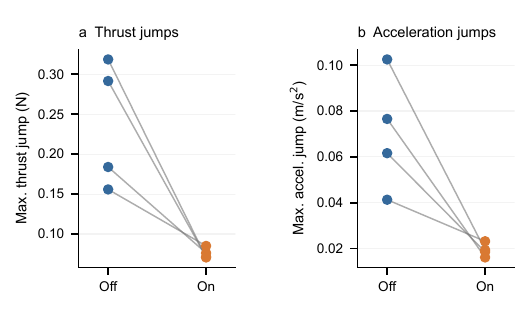}
\caption{Controlled thrust-difference penalty experiment with frozen network weights. Each line links the same original test layout with the penalty off and on; all four test layouts are included. Values are the maximum rotor-thrust or acceleration jump over both vehicles and the full common horizon. Two penalty-off layouts use nonlinear recovery after the initial six-step SCP attempt; all penalty-on layouts pass without that recovery.}
\label{fig:smoothness}
\end{figure}

Across the four original test layouts, the scene-averaged maximum rotor-thrust jump decreases from 0.23734 to 0.07581\,N, while the RMS jump decreases from 0.03802 to 0.02062\,N. The mean completion metric changes from 2.40005 to 2.40044\,s. Figure~\ref{fig:smoothness} includes every test-layout pair. The reported jumps are finite differences of piecewise-constant thrusts, which quantify discrete control variation.

The full pipeline permits nonlinear reinitialization followed by six SCP steps if the initial attempt fails. Two of the four penalty-off test layouts require this recovery, compared with none for penalty-on. Restricting the comparison to the two test layouts that both variants solve in the initial six steps still reduces the scene-averaged maximum thrust jump from 0.16980 to 0.08064\,N. This paired subset is reported separately from the four-layout full-pipeline result.

\subsection{Forward Runtime and Numerical Derivatives}
\begin{figure}[!t]
\centering
\includegraphics[width=\columnwidth]{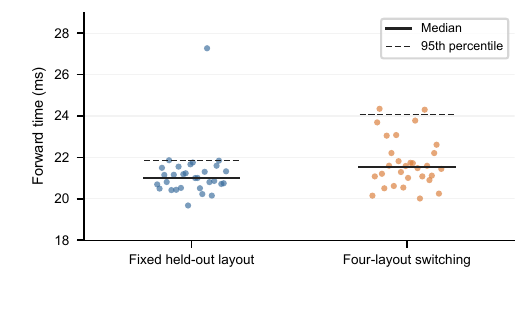}
\caption{All 60 recorded warm-reference forward timings, with 30 calls per setting. Solid horizontal marks show medians and dashed marks show 95th percentiles. Fixed-layout calls use the held-out exchange layout; switching includes two training, one validation, and one held-out layout. Each setting measures repeated solver calls with prepared trajectory references.}
\label{fig:runtime}
\end{figure}

The timed forward evaluation includes both networks, both local planners, and node-level feasibility checks. It excludes copying a stored reference into the planner and excludes the independent dense audit. Figure~\ref{fig:runtime} shows all recorded calls. Thirty repeated calls on the fixed held-out exchange layout have median 21.01\,ms and 95th percentile 21.86\,ms. Thirty calls switching across the four exchange layouts have median 21.54\,ms and 95th percentile 24.07\,ms. Every call passes the recorded feasibility and sampled separation checks. Both settings use matching numerical trajectory references and stop after one adaptive SCP round.

\begin{samepage}
The comparison of analytical and finite-difference derivatives, which is a local check of selected derivative entries, uses two selected final-layer weights on the exchange training layout B. The finite-difference perturbation is $10^{-4}$. The waypoint-network derivative is $2.11464\times10^{-4}$ analytically versus $2.15084\times10^{-4}$ numerically; the event-network derivative is $2.35052\times10^{-4}$ versus $2.33841\times10^{-4}$. Both pass the recorded tolerance of 5\% plus $10^{-7}$. Both perturbed plans remain feasible, and their event supports remain unchanged.
\par\end{samepage}

\FloatBarrier
\section{Discussion and Conclusion}
\begin{samepage}
The shared network adapts spatial waypoints and intermediate events so that independently solved rigid-body trajectories coordinate on a common physical grid. The final progress variable permits different completion metrics while retaining post-event motion. Thrust-difference regularization reduces the measured control jumps, and the forward backend provides millisecond-scale execution with matching trajectory references.
\par\end{samepage}

\begin{samepage}
The evidence concerns state-based two-vehicle planning with gates represented by position tolerance balls. Terminal CPC uses a local approximation, and inter-vehicle collision avoidance is learned through the outer objective and assessed by sampled trajectory audits. Within this scope, the implementation demonstrates event-conditioned coordination with nonlinear local planners and separately measured timing, feasibility, and sampled separation.
\par\end{samepage}

\end{document}